%% file: main_arxiv.tex
\documentclass[runningheads]{llncs}

\usepackage{eccvabbrv}

\usepackage{graphicx}
\usepackage{booktabs}
\usepackage{indentfirst}

\usepackage{tabularx}
\usepackage{booktabs}
\usepackage{array}
\usepackage{subcaption}
\usepackage{xcolor}

\usepackage[accsupp]{axessibility}  

\usepackage[hidelinks]{hyperref}
\usepackage[nameinlink,capitalize,noabbrev]{cleveref}

\usepackage{orcidlink}

\begin{document}

\title{RieMind: Geometry-Grounded Spatial Agent for Scene Understanding} 

\titlerunning{RieMind: Geometry-Grounded Spatial Agent for Scene Understanding}

\author{Fernando Ropero \and 
Erkin Turkoz \and 
Daniel Matos \and 
Junqing Du \and 
Antonio Ruiz \and 
Yanfeng Zhang\thanks{Corresponding author: zhangyanfeng8@huawei.com} \and 
Lu Liu \and 
Mingwei Sun \and 
Yongliang Wang}

\authorrunning{F. Ropero et al.}

\institute{Riemann Lab, Huawei Technologies}

\maketitle

\begin{figure}[ht]
  \centering
  \includegraphics[width=0.95\linewidth]{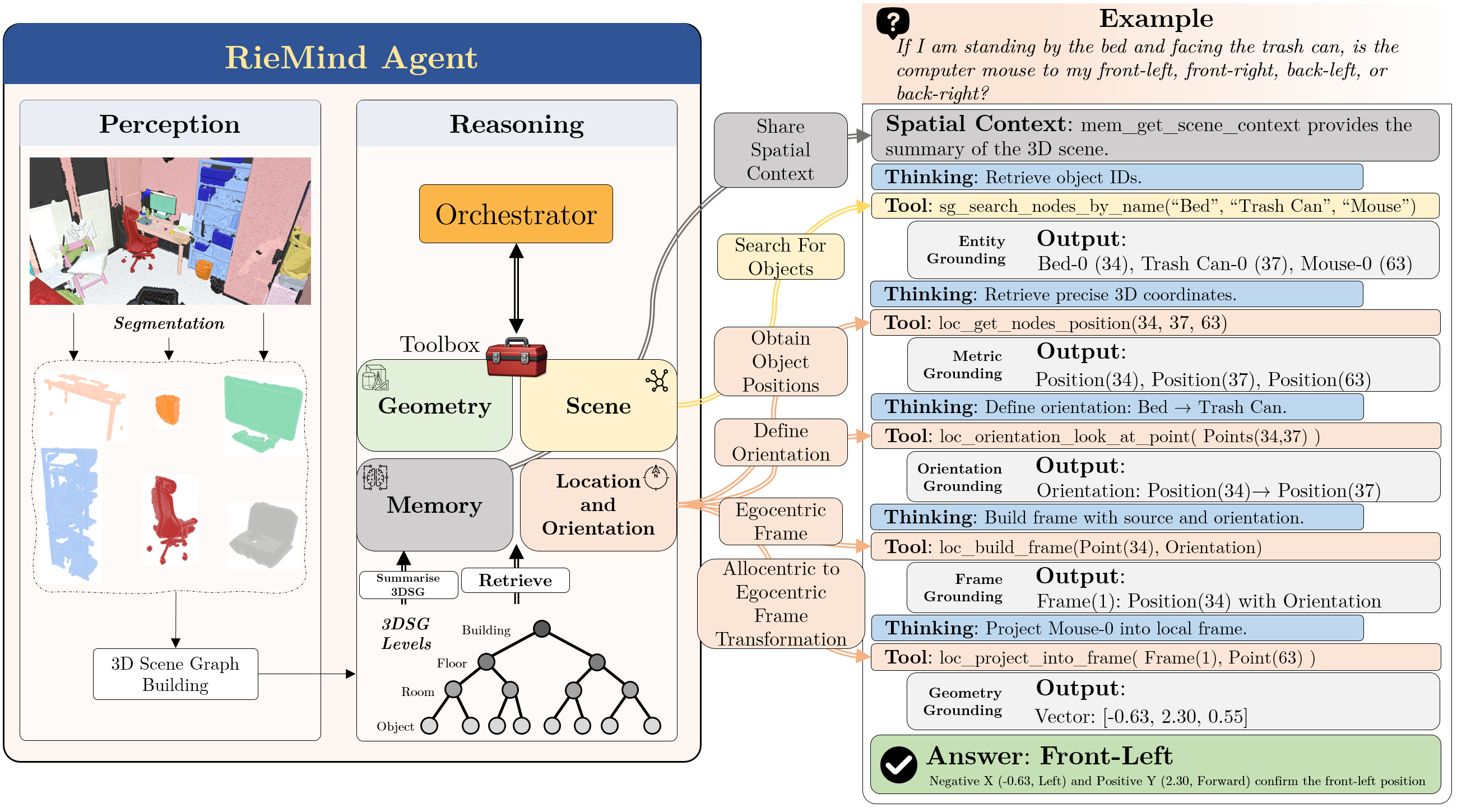}
  \caption{RieMind: Geometry-grounded framework for spatial agentic scene understanding, and in this figure, it is reasoning over the relative direction of objects in a scene, while performing allocentric-egocentric frame transformations.}
  \label{fig:spatial_reasoning_direction_example}
\end{figure}

\begin{abstract}
  Visual Language Models (VLMs) have increasingly become the main paradigm for understanding indoor scenes, but they still struggle with metric and spatial reasoning. Current approaches rely on end-to-end video understanding or large-scale spatial question answering fine-tuning, inherently coupling perception and reasoning. In this paper, we investigate whether decoupling perception and reasoning leads to improved spatial reasoning. We propose an agentic framework for static 3D indoor scene reasoning that grounds an LLM in an explicit 3D scene graph (3DSG). Rather than ingesting videos directly, each scene is represented as a persistent 3DSG constructed by a dedicated perception module. To isolate reasoning performance, we instantiate the 3DSG from ground-truth annotations. The agent interacts with the scene exclusively through structured geometric tools that expose fundamental properties such as object dimensions, distances, poses, and spatial relationships. The results we obtain on the static split of VSI-Bench provide an upper bound under ideal perceptual conditions on the spatial reasoning performance, and we find that it is significantly higher than previous works, by up to 16\%, without task specific fine-tuning. Compared to base VLMs, our agentic variant achieves significantly better performance, with average improvements between 33\% to 50\%. These findings indicate that explicit geometric grounding substantially improves spatial reasoning performance, and suggest that structured representations offer a compelling alternative to purely end-to-end visual reasoning.
\end{abstract}

\input{sections/introduction}

\input{sections/related_work}

\input{sections/methodology}

\input{sections/experiments}

\input{sections/conclusion}

\clearpage


%
%

\bibliographystyle{splncs04}
\bibliography{main}
\end{document}

%% file: sections/introduction.tex
\section{Introduction}\label{sec:intro}

The rapid rise of large language models (LLMs) and their linguistic and reasoning capabilities has left behind a reduced focus on spatial intelligence and understanding. This gap has then been filled by the development of visual language models (VLMs), that are able to concurrently understand visual information, and natural language. This multimodality has seen an ever increasing focus these past few years~\cite{10.1093/nsr/nwae403}, though some challenges remain concerning their spatial reasoning and understanding capabilities~\cite{Chen_2024_CVPR,stogiannidis2025mind}. These challenges become more severe when long-term visual and spatial context is required, such as long-form videos, showing a degradation in performance, leading to hallucinations~\cite{wang2024videohallucer,seth2025egoillusion}. Current VLMs lack robust, and grounded spatial understanding.

Spatial understanding is currently often gauged by question answering (QA) on either image or video benchmarks~\cite{cheng2024spatialrgpt,yang2025thinkingspacemultimodallarge}. These benchmarks range from 2D image relational spatial understanding~\cite{cheng2024spatialrgpt}, to 3D indoor spatial reasoning requiring metric information, such as distances, sizes, and geometries, usually utilising videos as inputs~\cite{yang2025thinkingspacemultimodallarge}. The two main approaches to indoor spatial understanding rely on, either large-scale spatial instruction fine-tuning of VLMs~\cite{feng2025visuospatialcognitiveassistant,ouyang2025spacerreinforcingmllmsvideo,li2025spatialladderprogressivetrainingspatial,wu2025reinforcingspatialreasoningvisionlanguage,tang2025videospatialreasoningobjectcentric,zhao2025spacemindcameraguidedmodalityfusion}, or tool-augmented reasoning~\cite{wu2025spatialscorecomprehensiveevaluationspatial,daxberger2025mmspatialexploring3dspatial}, where VLMs can reason by utilising geometric tools, such as depth, frames, bounding boxes, and others. However, both these approaches rely on a tight coupling between perception and implicit model representations, which can limit spatial metric consistency and interpretability. This is specifically important for 3D indoor environments, where reasoning should be grounded in persistent geometric representations.

This paper focuses on 3D indoor spatial understanding. While related works often frame this problem as a video understanding one, \ie~using a VLM to ingest videos and then apply them to QA benchmarks, we take a different conceptual approach. Indoor environments are often static in nature, hence we consider the input videos as a data ingestion and perception modality and not the basis for spatial reasoning. Therefore, we represent each scene using a 3D scene graph (3DSG) that is able to unify semantic, topological, and metric scene representations, providing a unified understanding of inter-object and intra-scene relationships~\cite{hughes2022hydra,werby2024hierarchical,takmaz2025search3d}. These 3DSGs are often used in robotics where precise geometric and spatial information are required for task planning, exploration, and object manipulation~\cite{saxena2024grapheqa,cao2025cognav,honerkamp2024language}. 

Building on this representation, we propose RieMind, an agentic framework that decouples perception from reasoning. Instead of using a VLM, we employ an LLM that invokes structured geometric tools and queries the 3DSG directly, retrieving explicit geometric information such as locations, distances, poses, dimensions, areas, volumes, and semantic relationships. The 3DSG is typically built from a separate perception module that performs instance segmentation followed by scene graph construction. In this paper, we assess an agent's spatial reasoning capabilities under ideal perceptual conditions by constructing the 3DSG from ground-truth annotations, eliminating errors introduced by upstream segmentation. This allows our framework to be agnostic to the upstream perception pipeline, as the scene information is abstracted and accessible exclusively through tool calls. We show that explicit geometric grounding yields improved spatial reasoning performance on VSI-Bench~\cite{yang2025thinkingspacemultimodallarge} when compared to previous approaches. The contributions of this paper are as follows:
\begin{itemize}
    \item We cast 3D indoor spatial understanding as a reasoning process over explicit geometric scene representations, by building an agentic, geometric, 3DSG-grounded framework that allows for interpretable metric and spatial reasoning for LLMs.
        
    \item We demonstrate state-of-the-art results on VSI-Bench, with an average increase of 16\% over competing fine-tuned spatial understanding models, and average improvements between 33\% and 50\% when comparing a base VLM with our agentic version.
\end{itemize}

%% file: sections/related_work.tex
\section{Related Work}\label{sec:related_work}

\subsubsection{VLMs for Indoor Spatial QA.} Indoor spatial understanding through QA has seen a surge of datasets and benchmarks, the most popular of which has been VSI-Bench~\cite{yang2025thinkingspacemultimodallarge}. Several works have improved the performance of base VLMs through spatial instruction QA fine-tuning or reinforcement learning, \eg ViCA~\cite{feng2025visuospatialcognitiveassistant}, SpaceR~\cite{ouyang2025spacerreinforcingmllmsvideo}, and others~\cite{li2025spatialladderprogressivetrainingspatial,wu2025reinforcingspatialreasoningvisionlanguage,tang2025videospatialreasoningobjectcentric,zhao2025spacemindcameraguidedmodalityfusion}. Other research directions include adding geometric priors or auxiliary representations into VLMs~\cite{zheng2025learningvideos3dworld,wu2025spatial}. Our work is somewhat complementary to this, as it is still focused on the same type of spatial understanding, but instead of the end-to-end implicit learning and reasoning, we allow for an LLM to ground its reasoning through tools that query an explicit geometric representation of a scene in the form of a 3DSG.

\subsubsection{Tool Integrated Spatial Reasoning.} A complementary research line focuses on augmenting VLMs with geometric tools to improve spatial reasoning, rather than fine-tuning directly, in order to reduce reliance on implicit priors and improve explanability of the reasoning process. This has been investigated in~\cite{wu2025spatialscorecomprehensiveevaluationspatial,han2025tigertoolintegratedgeometricreasoning,daxberger2025mmspatialexploring3dspatial}, and these works demonstrate that spatial understanding of VLMs benefits from access to geometric tools. The SpatialScore benchmark~\cite{wu2025spatialscorecomprehensiveevaluationspatial} introduces a tool-using agent, named SpatialAgent, with access to geometric and spatial tools. TIGeR~\cite{han2025tigertoolintegratedgeometricreasoning} applies a similar tool based agent but for robotic manipulation instead. MM-Spatial~\cite{daxberger2025mmspatialexploring3dspatial} also introduces tools, specifically depth. When it comes to tooling, these approaches have one thing in common, they rely on geometric estimation, as these tools estimate depth, pose, bounding boxes, and other quantities. This is where our works diverge, as we ground tool usage in a persistent 3DSG. We show that allowing the LLM to instead utilise tools that query the geometric information from a 3DSG improves performance significantly. We do this by assuming a given 3DSG, and create geometric tools from this structure. Effectively, we decouple perception from reasoning, unlike previous works.

\subsubsection{3DSG for Spatial Q\&A.} Prior work has developed 3DSGs as metric-semantic scene representations to encode explicitly indoor environments as nodes connected by semantic and spatial relationships~\cite{hughes2022hydra,werby2024hierarchical,takmaz2025search3d}. In the specific context of task planning, and exploration, 3DSGs have been utilised in embodied question answering (EQA)~\cite{saxena2024grapheqa,cao2025cognav,honerkamp2024language}, particularly for unexplored environments. These 3DSG QA applications differ quite significantly from our case with VSI-Bench, as we focus on spatial understanding, both absolute and relative. Furthermore, how a 3DSG is utilised in spatial QA also matters. These EQA works~\cite{saxena2024grapheqa,cao2025cognav,honerkamp2024language} supply the 3DSG in textual form within the model's prompt, while we allow the agent access to the 3DSG through a structured toolbox. Hence, 3DSGs have not been previously utilised on indoor 3D spatial understanding benchmarks, especially in such a systematic and structured way. In this work, we use the 3DSG as the basis for tool based spatial reasoning.

%% file: sections/methodology.tex
\section{Geometry-Grounded Agentic Framework}\label{sec:method}

Our framework decouples perception from spatial reasoning by enabling an LLM to access a set of geometric tools for spatial understanding. These geometric tools are grounded in the precise geometric and relational information stored in a 3D scene graph (3DSG). To isolate the spatial reasoning capabilities of the LLM, the 3DSG is built using ground truth annotations, allowing to understand the upper bound on their spatial reasoning performance. Our framework consists of two components: a perception layer that constructs a 3DSG, and an agentic reasoning layer that interacts with our geometry-grounded toolbox.

\subsection{Perception Layer}\label{sec:perception_layer}

On the perception aspect of this framework, we need to construct the 3DSG. For the construction of the 3DSG, we follow the procedure in~\cite{werby2024hierarchical}, where typically the input to build the 3DSG is a set of RGB-D frames and associated camera parameters. This 3DSG can be formalised as 

\begin{equation}
\mathcal{G} = (\mathcal{N}, \mathcal{E}),
\end{equation}
where $\mathcal{N}$ describes nodes, and $\mathcal{E}$ describes edges. The set of nodes can be expressed as the union of the four different node layers of the graph, given by a building node $\mathcal{N}_{B}$, floors $\mathcal{N}_{F}$, rooms $\mathcal{N}_{R}$, and object $\mathcal{N}_{O}$. The set of edges $\mathcal{E}$ is given by the union of the edges between the nodes at different layers, as well as between object nodes. Concretely, $\mathcal{E}_{BF}$ denotes the edges between the building and floor nodes, $\mathcal{E}_{FR}$ between the floor and rooms, $\mathcal{E}_{RO}$ between the rooms and object, and finally, $\mathcal{E}_{OO}$ describes the inter-object relational edges. Besides this hierarchical connectivity, the 3DSG is a metric representation, as it contains metric information for every node, and between them as well if desired. Each of these nodes stores additional attribute information. The building node $\mathcal{N}_{B}$ describes the entire scene, and stores class attributes. The floor nodes $\mathcal{N}_{F}$ group rooms belonging to the same floor, and store basic geometric information, such as areas, and bounding box based geometric information. Room nodes $\mathcal{N}_{R}$ represent a bounded spatial region corresponding to a room, hence they will only store bounding box related and derived geometric information. Each object node $\mathcal{N}_{O}$ stores semantic information, from a closed-set of classes, and metric information such as bounding box dimensions, volume, surface area, orientation, and location. Finally, inter-object edges $\mathcal{E}_{OO}$ can optionally store relation labels when available. While it is easy to derive topological and geometric relationships such as $\{\textit{contains}, \textit{supports}, \textit{near}, \textit{under}\}$, we restrict inter-object edges to $\{\textit{near}\}$. This relation alone is sufficient for an agent to reason on the additional topological and geometric relationships when querying location related information. Each node in $\mathcal{N}$ is identified by a unique node ID, that is persistent to a scene, and serves as a linking mechanism between the 3DSG and the majority of tool calls.

\subsection{Reasoning Layer}\label{sec:reasoning_layer}

\subsubsection{Agentic Architecture and Prompt.}\label{sec:framework_and_prompt}

\begin{figure}[ht]
  \centering
  \includegraphics[width=0.95\linewidth]{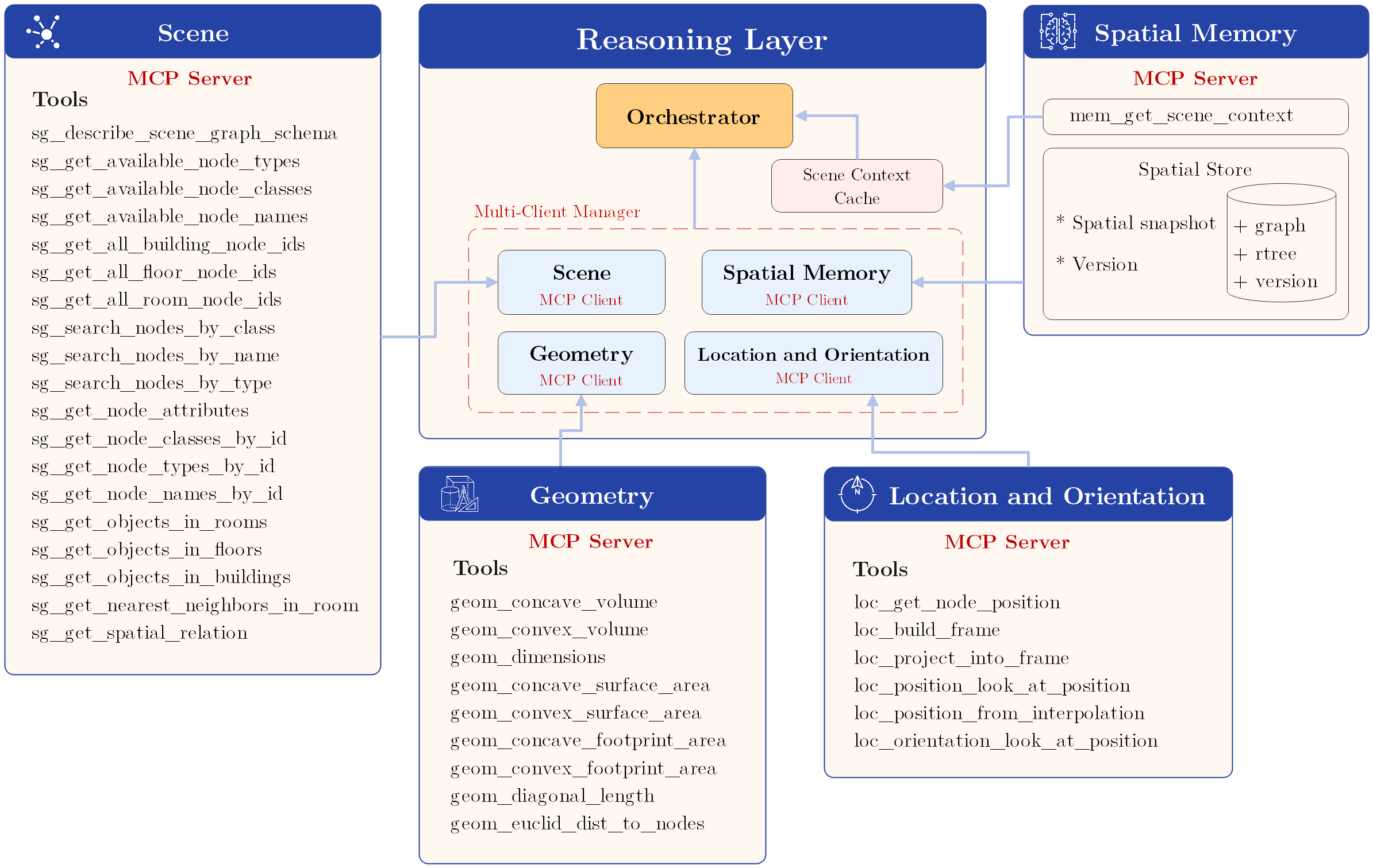}
  \caption{Reasoning component of the spatial agent framework, containing the different types of tools, and general agent architecture following the MCP.}
  \label{fig:tool_framework}
\end{figure}

The architecture is organised through distributed Model Context Protocol (MCP) servers that expose our set of tools to the agent, and they are grouped into semantic namespaces, \ie~\textit{memory}, \textit{scene}, \textit{geometry}, and \textit{location and orientation}. This separation conceptually limits the types of tool calls teh agent can perform, and builds an immediate distinction between the searching for objects, their geometries, and their orientations. This semantic tool distinction also facilitates the traceability of the reasoning process of the agent. \cref{fig:tool_framework} displays the general framework and tool calling system that we utilise, where every tool call is rooted in the 3DSG.

The prompting of the agent is structured in two components: a system prompt, and the user question. The system prompt is divided into several sections: 
\begin{enumerate}
  \item Role definition establishing the agent as a spatial reasoning system that must delegate all geometric computations to tools, without performing calculations internally.
  \item List of available tools in a static schema.
  \item Scene context cache obtained from the \textit{mem\_get\_scene\_context} tool. This tool sits in the \textit{memory} semantic namespace, which provides the summary of the 3DSG of that particular scene, \ie~the complete 3DSG hierarchy and the objects within, allowing the agent to understand the content of the scene it is currently in. Please refer to \cref{table:scene_graph_context} as an example that can be provided to the agent.
  \item Core agent behaviour enforcing six constraints: resolve user terminology against the scene graph before querying; search for referenced objects before computations; avoid duplicate tool calls; enforce valid argument types; issue one tool call per response; and maintain an explicit reasoning plan updated after each step.
  \item Tool data flow defines the general structure required to use the tools, \ie~object names need to be disambiguated from the question, resolve them to the scene graph classes, and obtain node identifiers before calling the tools.
  \item Tool catalog is the reference documentation for the four namespaces: sg\_* (scene graph discovery and structural queries), geom\_* (geometric measurements including dimensions, volumes, areas, and distances), mem\_* (scene context retrieval), and loc\_* (spatial reference frame construction and projection). This includes function arguments and outputs.
  \item Answer format, requiring a JSON output schema with three fields: a natural language summary, evidence of tools invoked, and data dictionary of key results.
\end{enumerate}
The user questions can be any questions about the scene, but here, we focuson the questions from VSI-Bench~\cite{yang2025thinkingspacemultimodallarge}. The only parts that can vary are the user question, and the scene context summary that will change if a particular question refers to a different scene.

\begin{table}[t]
\centering
\caption{Example of the scene graph context provided to the agent that is integrated into the system prompt. This scene context is going to be dependent on the actual scene, as each will contain different contents and structure.}
\label{table:scene_graph_context}
\begin{tabular}{c c c}
\toprule
Node Type & Name / Identifier & Class (Count) \\ 
\midrule
BuildingNode & Building-0 & residential \\
FloorNode & Floor-0 & 0 \\
RoomNode & Room-0 & room\_0 \\
\midrule
ObjectNode & Cabinet-0 \ldots Cabinet-15 & cabinet (16) \\
ObjectNode & Chair-0 \ldots Chair-2 & chair (3) \\
ObjectNode & Oven-0, Oven-1 & oven (2) \\
ObjectNode & Shelf-0 & shelf (1) \\
ObjectNode & Sink-0 & sink (1) \\
ObjectNode & Sofa-0, Sofa-1 & sofa (2) \\
ObjectNode & Stool-0 & stool (1) \\
ObjectNode & Stove-0 & stove (1) \\
ObjectNode & Table-0, Table-1 & table (2) \\
ObjectNode & Tv Monitor-0 & tv\_monitor (1) \\
\midrule
\multicolumn{3}{l}{Total: 30 ObjectNodes, 1 RoomNode, 1 FloorNode, 1 BuildingNode}
\\
\bottomrule
\end{tabular}
\end{table}

\subsubsection{Geometry-Grounded Toolbox.}\label{sec:tools}

\begin{figure}[ht]
  \centering
  \includegraphics[width=0.95\linewidth]{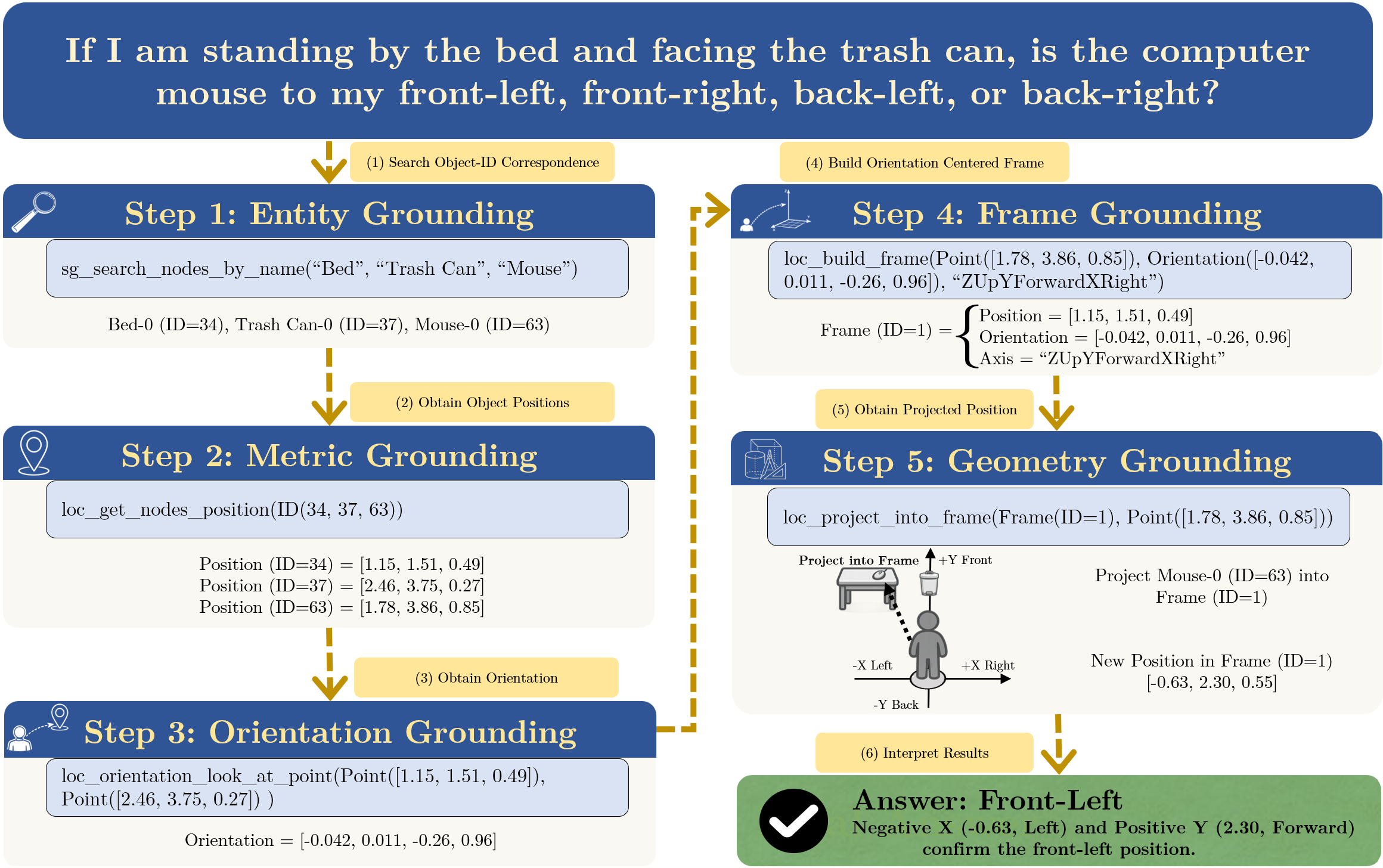}
  \caption{Exemplifying the different types of explicit grounding through a relative direction question. For any question posed to the agent, it will need to disambiguate the actual entity, and this grounds its reasoning. For this question, it then needs to obtain positions, orientations, and create the specific egocentric frame with its own axis convention. It can then perform the projection of the desired object into this new egocentric frame. All these tools and processes ground the reasoning sequence of the agent, and force it to stay focused on these basic geometric objects.}
  \label{fig:reasoning_grounding}
\end{figure}

To enable spatial reasoning from natural language alone, we create a comprehensive tool framework to enable the agent to reason over a deterministic data structure in the 3DSG. Rather than implicitly encoding the metric and relational scene objects and information, we do so explicitly. Our tool framework is constructed according to three design principles:

\begin{itemize}
  \item \textbf{Minimal geometric primitives.} Tools implement atomic geometric or structural operations rather than composite functions or reasoning shortcuts. Each tool either access only one attribute from the object, or performs a single mathematical transformation, \eg~accessing the volume of an object or projecting a chosen object into a known frame.
  \item \textbf{Explicit grounding.} Each node in the 3DSG has a unique persistent ID, every geometry and orientation tools operate on node IDs rather than free-form text, enforcing referential consistency and entity, with free-from text only being permitted in the 3DSG search tools. This structure allows the agent to initially reason over the semantic information, and then to query the explicit ID referenced object, aiding consistency across reasoning steps. In a specific scene there are often several instances of the same object, and to reason correctly the agent is helped by the entity grounding through IDs. In \cref{fig:reasoning_grounding}, we display the different types of grounding that aid the reasoning process of the agent, the entity, metric, orientation, frame, and geometric grounding through the 3DSG information.
  \item \textbf{Determinism.} Tool outputs depend solely on the 3DSG state and are independent of the agent's internal reasoning.
\end{itemize}

In addition to these design principles, to faciliate the agent's reasoning process, we divide our framework into four semantic categories, so that choosing a particular tool becomes easier. These categories are given by \textit{memory}, \textit{scene}, \textit{geometry}, and \textit{location and orientation}:

\begin{itemize}
\item \textbf{Memory Tool} provides a structured summary of the current 3DSG, allowing the agent to understand the environment it needs to query.

\item \textbf{Scene Tools} provide direct query access to the nodes (buildings, floors, rooms, objects) and edges of the scene graph, either through input strings, ids, or even by distance. These tools allow for entity grounding, scope restriction, and nearest neighbour node searches. For example, they allow to retrieve all objects within a room, or of a given class, and even the semantic relationships between such objects.

\item \textbf{Geometric Tools} expose precise metric properties derived from each of the node's 3D geometry. This includes bounding box size, volume, surface area, and distance computations, with explicit support for both convex and concave geometry to handle irregular indoor objects. These tools enable explicit geometric understanding and reasoning allowing for quantitative comparisons that would be difficult with visual features only.

\item \textbf{Location and Orientation Tools} allow for egocentric-allocentric transformation reasoning, which is the basis of VSI-Bench~\cite{yang2025thinkingspacemultimodallarge}. In essence, these tools allow for explicit access to reference frames and user poses. They support constructing and manipulating coordinate frames, and projecting object frames into an egocentric one, and vice-versa. These tools are crucial in resolving pose dependent questions, \eg~if something is to the right of the user when facing a specific object.
\end{itemize}
This extensive list of tools can be seen in \cref{fig:tool_framework}. They are deterministic, and are as close to the basic geometries as possible, in order to describe these static environments. They let an agent perform a sequence of operations to understand symbolic, geometric, and frame dependent questions about the environment in an auditable and interpretable way. Creating a toolbox that only contains targeted functions that expose basic geometric information of the components of a scene allows to understand if an agent is able to reason from the fundamental geometry itself, while decoupling perception from reasoning. This toolbox does not contain general functions that are a composition of more specific ones, and this is a crucial component of our framework.

%% file: sections/experiments.tex
\section{Experiments}\label{sec:experiments}

We evaluate our geometry-grounded agent on the static spatial questions of VSI-Bench~\cite{yang2025thinkingspacemultimodallarge}, which measures 3D indoor spatial reasoning through around 5000 QA pairs, derived from 288 real videos. These videos were obtained from the validation sets of three 3D scene datasets, ARKitScenes~\cite{baruch2022arkitscenesdiverserealworlddataset}, ScanNet~\cite{dai2017scannet}, and ScanNet++~\cite{yeshwanthliu2023scannetpp}. For this static portion of VSI-Bench, we discard the \textit{route planning}, and \textit{order appearance} question which are fundamentally dynamic in nature. From around 5000 original questions, this leaves 4185 questions in the remaining six question types that we evaluate here. To isolate the geometric reasoning from the perception systems, and obtain a rough upper bound for LLM performance on VSI-Bench, we build the 3DSG from ground truth annotations in the manner specificied in \cref{sec:method}. This experimental setup follows prior work that evaluates spatial reasoning performance using ground truth annotations, e.g. depth~\cite{daxberger2025mmspatialexploring3dspatial}, and this should be understood as measuring spatial reasoning rather than end-to-end perception and reasoning. Effectively, we are measuring a reasoning upper bound, given ground truth data which is certainly of high quality but still demonstrates annotation mistakes, omissions, and occlusions. With our framework, we evaluate the agentic performance of Qwen2.5-VL-7B~\cite{bai2025qwen25vltechnicalreport}, GPT-4o~\cite{openai2024gpt4ocard}, and GPT-4.1.

\subsection{Impact of Agentic Reasoning}\label{sec:results_agent_based}

\cref{table:vlm_base_vs_tool}~demonstrates the performance increase on VSI-Bench of base VLMs, \ie~without any access to tools or without fine-tuning, when compared with their agentic version. The large improvements indicate that questions from the absolute categories such as counting, absolute sizes and distances, can be largely solved through this 3DSG geometric grounding, as it provides a source of geometric truth that the agent can access and reason over. The performance on the \textit{relative direction} and \textit{relative distance} questions, demonstrate that this is still reasoning dependent, hence the substantial improvements with GPT-4o, but the modest to no improvements with Qwen2.5-VL-7B, as it is a smaller model with worse reasoning capabilities~\cite{bai2025qwen25vltechnicalreport,guan2024hallusionbenchadvanceddiagnosticsuite}.

\begin{table}[t]
\centering
\caption{Comparison of performance on VSI-Bench between base VLMs (without tools), and their LLM counterpart as an agent with our tool framework. Improvements are shown in parentheses next to the Agent + Tools scores.}
\label{table:vlm_base_vs_tool}
\begin{tabular}{l|l @{\hspace{10pt}} l|l @{\hspace{10pt}} l}
\toprule
& \multicolumn{2}{c|}{Qwen2.5-VL-7B~\cite{bai2025qwen25vltechnicalreport}} & \multicolumn{2}{c}{GPT-4o~\cite{openai2024gpt4ocard}} \\
\cmidrule(lr){2-3}\cmidrule(lr){4-5}
Question Type & \textit{Base} & \textit{Agent + Tools} & \textit{Base} & \textit{Agent + Tools} \\

\midrule

Object Count  & 40.9 & \textbf{89.7} (+48.8) & 46.2 & \textbf{85.1} (+38.9) \\
Absolute Distance  & 14.8 & \textbf{90.3} (+75.5) &  5.3 & \textbf{93.2} (+87.9) \\
Object Size   & 43.4 & \textbf{93.6} (+50.2) & 43.8 & \textbf{96.5} (+52.7) \\
Room Size   & 10.7 & \textbf{31.9} (+21.2) & 38.2 & \textbf{83.5} (+45.3) \\
Relative Distance  & 38.6 & \textbf{44.5} (+5.9)  & 37.0 & \textbf{85.6} (+48.6) \\
Relative Direction   & \textbf{38.5} & 34.7 \phantom{.}(-3.8)     & 41.3 & \textbf{67.5} (+26.2) \\

\midrule

Average        & 31.2 & \textbf{64.1} (+32.9) & 35.3 & \textbf{85.2} (+49.9) \\

\bottomrule
\end{tabular}
\end{table}

The improvements on \cref{table:vlm_base_vs_tool}~are related to the compositional complexity of each of the VSI-Bench questions. Questions related to object counting, absolute distance, object size estimation, room size estimation are rather simple in tool calling sequences and rely on one or two tool invocations, see \cref{table:tool_complexity}. The compositional nature of the relative direction questions that require deep reasoning chains (5-6 tool calls), cause more frequent hallucinations, particularly for smaller models. Hence, the degrading performance of Qwen2.5-VL-7B on these questions as seen in \cref{table:vlm_base_vs_tool}. Despite these issues, \cref{table:vlm_base_vs_tool}~demonstrates that our specific tool framework and usage, grounded in the 3DSG, allows for much better spatial reasoning than base VLMs. Considering that the largest increases were seen on GPT-4o, we can infer that these base geometric tools enable larger models to perform more complex reasoning.

\begin{table}[t]
\centering
\caption{Score vs.\ reasoning complexity by question type for Qwen2.5-VL-7B on VSI-Bench. Relative direction questions require a 5-tool pipeline (\texttt{sg\_search} $\to$ \texttt{loc\_get\_position} $\to$ \texttt{loc\_orientation} $\to$ \texttt{loc\_build\_frame} $\to$ \texttt{loc\_project}), while simpler types need only 1--2 tools. Qwen2.5-VL-7B frequently fails to complete the full pipeline, resulting in lower accuracy.}
\label{table:tool_complexity}
\small
\setlength{\tabcolsep}{9pt} 
\begin{tabular}{lcccc}
\toprule
Question Type & Number of Questions & Score & \multicolumn{2}{c}{Tools} \\
\cmidrule(lr){4-5}
& & & Average & Median \\
\midrule
Object Count        & 544 & 89.7 & 1.05 & 1 \\
Absolute\ Distance  & 799 & 90.3 & 2.14 & 2 \\
Object Size       & 927 & 93.6 & 2.15 & 2 \\
\midrule
Room Size        & 282 & 31.9 & 2.35 & 2 \\
Relative\ Distance         & 703 & 44.5 & 2.28 & 2 \\
\midrule
Relative\ Direction (Easy) & 211 & 46.1 & 4.06 & 4 \\
Relative\ Direction (Med.) & 360 & 31.7 & 3.83 & 4 \\
Relative\ Direction (Hard) & 359 & 31.1 & 3.89 & 4 \\
\bottomrule
\end{tabular}
\end{table}

\subsection{Upper Bound Evaluation on VSI-Bench}\label{sec:vsi_bench_results}

\cref{table:vsi_bench}~demonstrates that our geometry grounded agentic framework achieves state-of-the-art performance on the static questions of VSI-Bench, surpassing proprietary, open-source, and fine-tuned models without the need for additional task-specific training. These results should be interpreted as near upper bounds on the agent's VSI-Bench performance, as we focus only on the reasoning capabilities over geometric tools. Nevertheless, the performance increase from the agentic GPT4.1 when compared to both other proprietary and fine-tuned models is significant, from the closest average results of 73.6 from SpaceMind~\cite{zhao2025spacemindcameraguidedmodalityfusion} to 89.5 with GPT-4.1. The gap from GPT-4o to GPT-4.1 is particularly important, as the most complicated static questions are those of relative direction, and in these questions the improved reasoning capabilities of GPT-4.1 allow it to increase almost 20 points from GPT-4o. This supports the fact that the performance on these static spatial questions, and with our agentic framework and toolbox, relies mostly on the reasoning capabilities of the model. We also note that the average performance of Qwen2.5-VL-7B (64.1) is higher than most of the fine-tuned models displayed here, trailing only VLM-3R~\cite{fan2025vlm3rvisionlanguagemodelsaugmented}, and SpaceMind~\cite{zhao2025spacemindcameraguidedmodalityfusion}, and this is without any scene-specific training, which speaks to the potential generality of our framework. Another advantange of our framework is that it does not rely on any further training, unlike the fine-tuned models in~\cref{table:vsi_bench}, which typically require significant computational resources. We argue that due to the substantial differences in performance between our agents and fine-tuned models in the absolute questions, such as object counting, absolute distance, and size estimation, that metric estimation and reasoning is better served with geometric grounding rather than by learned visual priors.

\begin{table*}[t]
\centering
\caption{Evaluation on the static questions of VSI-Bench, the comparison includes proprietary, open-source, and fine-tuned models. Our agentic framework achieves state-of-the-art performance across all subtasks, where bold text denotes best performance, and underline denotes second best. These results should still be understood as an upper-bound on spatial understanding performance.}\label{table:vsi_bench}
\resizebox{\textwidth}{!}{
\begin{tabular}{c|c|c|c|c|c|c|c}
\toprule
Models & Avg. &  Obj. Count & Abs. Dist. & Obj. Size & Room Size & Rel. Dist. & Rel. Dir. \\
\midrule
\multicolumn{8}{l}{\textit{Proprietary}} \\
GPT-4o & 35.3 & 46.2 & 5.3 & 43.8 & 38.2 & 37.0 & 41.3 \\
Gemini-1.5 Flash & 44.5 & 49.8 & 30.8 & 53.5 & 54.4 & 37.7 & 41.0 \\
Gemini-1.5 Pro & 48.7 & 56.2 & 30.9 & 64.1 & 43.6 & 51.3 & 46.3 \\
\midrule
\multicolumn{8}{l}{\textit{Open-Source}} \\
InternVL3-78B & 50.7 & 71.2 & 53.7 & 44.4 & 39.5 & 55.9 & 39.5 \\
LLaVA-NeXT-Video-7B & 36.7 & 48.5 & 14.0 & 47.8 & 24.2 & 43.5 & 42.4 \\
LLaVA-NeXT-Video-72B & 40.6 & 48.9 & 22.8 & 57.4 & 35.3 & 42.4 & 36.7 \\
Qwen2.5-VL-7B & 31.2 & 40.9 & 14.8 & 43.4 & 10.7 & 38.6 & 38.5 \\
LLaVA-OneVision-7B & 34.2 & 47.7 & 20.2 & 47.4 & 12.3 & 42.5 & 35.2 \\
LLaVA-OneVision-72B & 40.8 & 43.5 & 23.9 & 57.6 & 37.5 & 42.5 & 39.9 \\
\midrule
\multicolumn{8}{l}{\textit{Fine-tuned}} \\
SpaceR~\cite{ouyang2025spacerreinforcingmllmsvideo} & 46.4 & 57.8 & 28.2 & 59.9 & 47.1 & 40.1 & 45.4 \\
VG-LLM~\cite{zheng2025learningvideos3dworld} & 52.3 & 67.9 & 37.7 & 58.6 & 62.0 & 46.6 & 40.7 \\
ViLaSR~\cite{wu2025reinforcingspatialreasoningvisionlanguage} & 47.3 & 63.5 & 34.4 & 60.6 & 30.9 & 48.9 & 45.2 \\
Spatial-MLLM~\cite{wu2025spatial} & 49.3 & 65.3 & 34.8 & 63.1 & 45.1 & 41.3 & 46.2 \\
VLM-3R~\cite{fan2025vlm3rvisionlanguagemodelsaugmented} & 67.0 & 70.2 & 49.4 & 69.2 & 67.1 & 65.4 & 80.5 \\
OCR~\cite{tang2025videospatialreasoningobjectcentric} & 47.8 & 63.2 & 34.1 & 57.4 & 46.7 & 39.6 & 45.5 \\
ViCA~\cite{feng2025visuospatialcognitiveassistant} & 63.5 & 68.8 & 57.0 & 79.2 & 75.1 & 58.5 & 42.6 \\
SpaceMind~\cite{zhao2025spacemindcameraguidedmodalityfusion} & 73.6 & 73.3 & 61.4 & 77.3 & 74.2 & 67.2 & \textbf{88.4} \\

\midrule
\multicolumn{8}{l}{\textit{Ours}} \\

Qwen2.5-VL-7B & 64.1 & \textbf{89.7} & 90.3 & 93.6 & 31.9 & 44.5 & 34.7 \\
GPT-4o & \underline{85.2} & 85.1 & \underline{93.2} & \underline{96.5} & \textbf{83.5} & \underline{85.6} & 67.5 \\
GPT-4.1 & \textbf{89.5} & \underline{86.5} & \textbf{94.9} & \textbf{97.9} & \underline{77.8} & \textbf{92.7} & \underline{87.3} \\

\bottomrule
\end{tabular}
}
\end{table*}

%% file: sections/conclusion.tex
\section{Conclusion}\label{sec:conclusion}

We present a geometry-grounded agentic framework for 3D indoor spatial understanding, by constructing a 3DSG that the LLM can query through geometry-based tools, effectively decoupling perception from reasoning. We evaluate on the static questions of VSI-Bench with a 3DSG instantiated from ground truth annotations, allowing results to be interpreted as an approximate upper bound on spatial reasoning performance. Nevertheless, our results demonstrate that explicitly grounding spatial reasoning through geometric primitives yields strong performance, particularly for metric tasks such as distance and size estimation. Furthermore, our agentic results show an increased performance of 33\% for Qwen2.5-VL-7B, and 50\% for GPT-4.1, over the base VLM results; and an 16\% increase over the latest fine-tuned spatial QA VLMs (SpaceMind~\cite{zhao2025spacemindcameraguidedmodalityfusion}) without any further training. Despite this, our results suggest that reasoning capability is the primary bottleneck for compositionally complex questions such as relative direction, with smaller models performing worse. The primary limitation of this work is the reliance on ground truth annotations, and constructing a 3DSG from RGB-D data remains the natural next step. This framework can potentially also be applied to VLMs in an attempt to geometry-ground their visual reasoning process.